# Soft Diamond Regularizers for Deep Learning

Olaoluwa Adigun[a] and Bart Kosko[b]*

[a]*Knight Foundation School of Computing and Information Sciences, Florida International University, Miami, USA.*

[b]*Department of Electrical and Computer Engineering, University of Southern California, Los Angeles, USA.*

## Abstract

This chapter presents the new family of *soft diamond* synaptic regularizers based on thick-tailed symmetric alpha stable ($\mathcal{S}\alpha\mathcal{S}$) probability bell curves. These new parametrized weight priors improved deep-learning performance on image and language-translation test sets and increased the sparsity of the trained weights. They specifically outperformed the state-of-the-art hard-diamond Laplacian regularizer of sparse lasso regression and classification. The $\mathcal{S}\alpha\mathcal{S}$ synaptic weight priors have power-law bell-curve tails that are thicker than the thin exponential tails of Gaussian bell curves that underly ridge regularizers. Their tails get thicker as the $\alpha$ parameter decreases. These thicker tails model more impulsive behavior and allow for occasional distant search in synaptic weight spaces of extremely high dimension. The geometry of their constraint sets has a diamond shape. The shape varies from a circle to a star or diamond that depends on the $\mathcal{S}\alpha\mathcal{S}$ tail thickness and dispersion of the $\mathcal{S}\alpha\mathcal{S}$ weight prior. These $\mathcal{S}\alpha\mathcal{S}$ bell curves lack a closed form in general and this makes direct training computationally intensive. We removed this computational bottleneck by using a precomputed look-up table. We tested the soft diamond regularizers with deep neural classifiers on both image test sets and German-to-English language translation. The best regularizer in almost all cases was the soft diamond with the very thick-tailed sub-Cauchy value $\alpha = 0.5$. The image simulations used the three datasets CIFAR-10, CIFAR-100, and Caltech-256. The regularizers improved the accuracy and sparsity of the classifiers. The classification gains included 4.57% on CIFAR-10, 4.27% on CIFAR-100, and 6.69% on Caltech-256. We also tested with deep neural machine-translation models on the IWSLT-2016 Evaluation dataset for German-to-English text translation. The gains with neural machine translation models included a 5.78% increase in BLEU score, a 5.73% increase in METEOR score, and a 7.98% increase in the F1 score for the ROUGE-L metric. They also outperformed $\ell_2$ or Gaussian regularizers on all the test cases. Soft diamond regularizers further outperformed $\ell_1$ lasso or Laplace regularizers because they had greater sparsity while still improving classification accuracy. These findings recommend the sub-Cauchy $\alpha = 0.5$ soft diamond regularizer as a competitive and sparse regularizer for large-scale machine learning.

## Section I: Soft-diamond Regularizers from Thick-tailed Synaptic Priors

This chapter explores deep learning with the new soft-diamond synaptic regularizers in the Bayesian framework of thick-tailed symmetric alpha-stable $S\alpha S$ bell-curve weight priors. These parametrized bell curves offer a rich and efficient family of alternatives to standard ridge (Gaussian prior) regularizers and lasso (Laplacian prior) regularizers. Ridge regressors [1], [2] form a spherical constraint shape while lasso regularizers [3], [4] form a hard diamond.

Figure 1 shows the soft diamond that performed best in almost all simulations that compared uniform, Gaussian, and Laplace priors with those based on $S\alpha S$ priors where the alpha ranged from 0.1 to 1.9. This best soft diamond had an extremely thick bell-curve tail of $\alpha = 0.5$. It modeled highly impulsive search events in synaptic weight space whereas a thin-tailed Gaussian prior with $\alpha = 2$ searches only locally. Such high impulsivity appeared to allow this soft diamond prior to balance the benefits of local search with the benefits of an occasional but quite distant search in synaptic spaces of extremely high dimension.

Figure 2 shows 4 types of $S\alpha S$ bell curves and their plots of white noise where the thin-tailed Gaussian has the highest alpha value of $\alpha = 2$ and where it alone has finite variance. The Cauchy bell curve corresponds to the thick-tailed case of $\alpha = 1$. The new family of stable-bell-curve regularizers gives a constraint set with a soft-diamond shape as in Figure 3. These new S$\alpha$S weight priors combine with the stochastic backpropagation (BP) algorithm to train deep neural classifiers with sparse sets of trained weights between hidden layers. Table 1 shows that these soft-diamond priors outperformed ridge and lasso regularizers on CIFAR-10, CIFAR-100, and Caltech-256. Table 2 shows that these new $S\alpha S$ priors further improved classification accuracy on CIFAR-10 when combined with dropout, batch, or data-augmentation regularizers. Tables 3-4 show that these new $S\alpha S$ priors also improved the performance of deep neural language translation models.

We first cast BP as maximum-likelihood estimation of synaptic parameters and then extend it to the Bayesian case of posterior maximization given weight priors.

We start with the simpler case of maximum likelihood. Then BP training seeks the parameter vector or array $\theta^*$ that maximizes the neural network's forward likelihood $p(y|x,\theta)$:

$$\theta^* = \arg\max_{\theta}\; p(y|x,\theta) = \arg\max_{\theta} \ln p(y|x,\theta)\ . \tag{1}$$

This maximization equivalently minimizes the cross-entropy between the target vector and the classifier's output activation vector $a^y$. BP training iteratively updates the synaptic weights as it

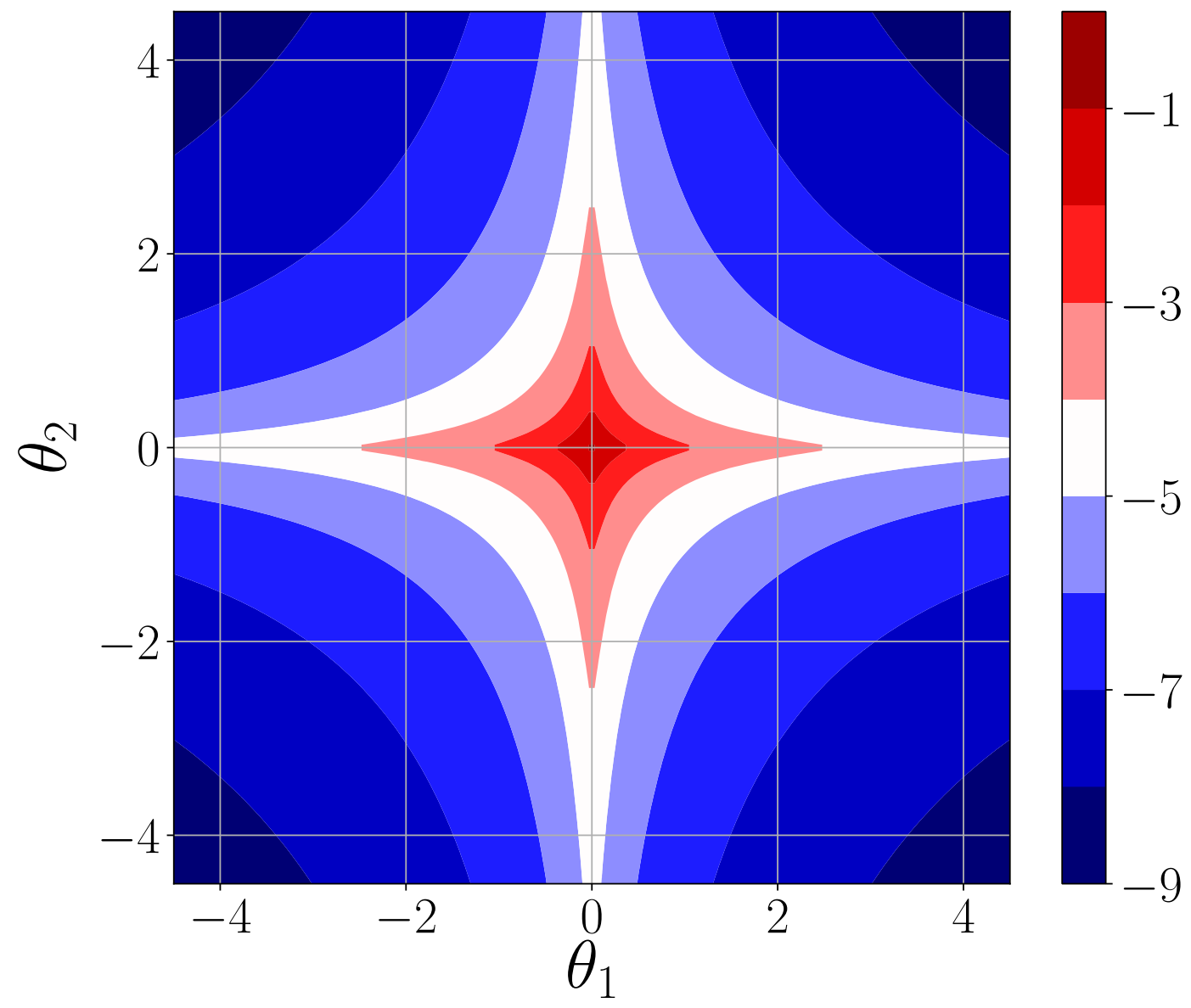

(a) Sub-Cauchy Soft Diamond with thick $\alpha = 0.5$ tail

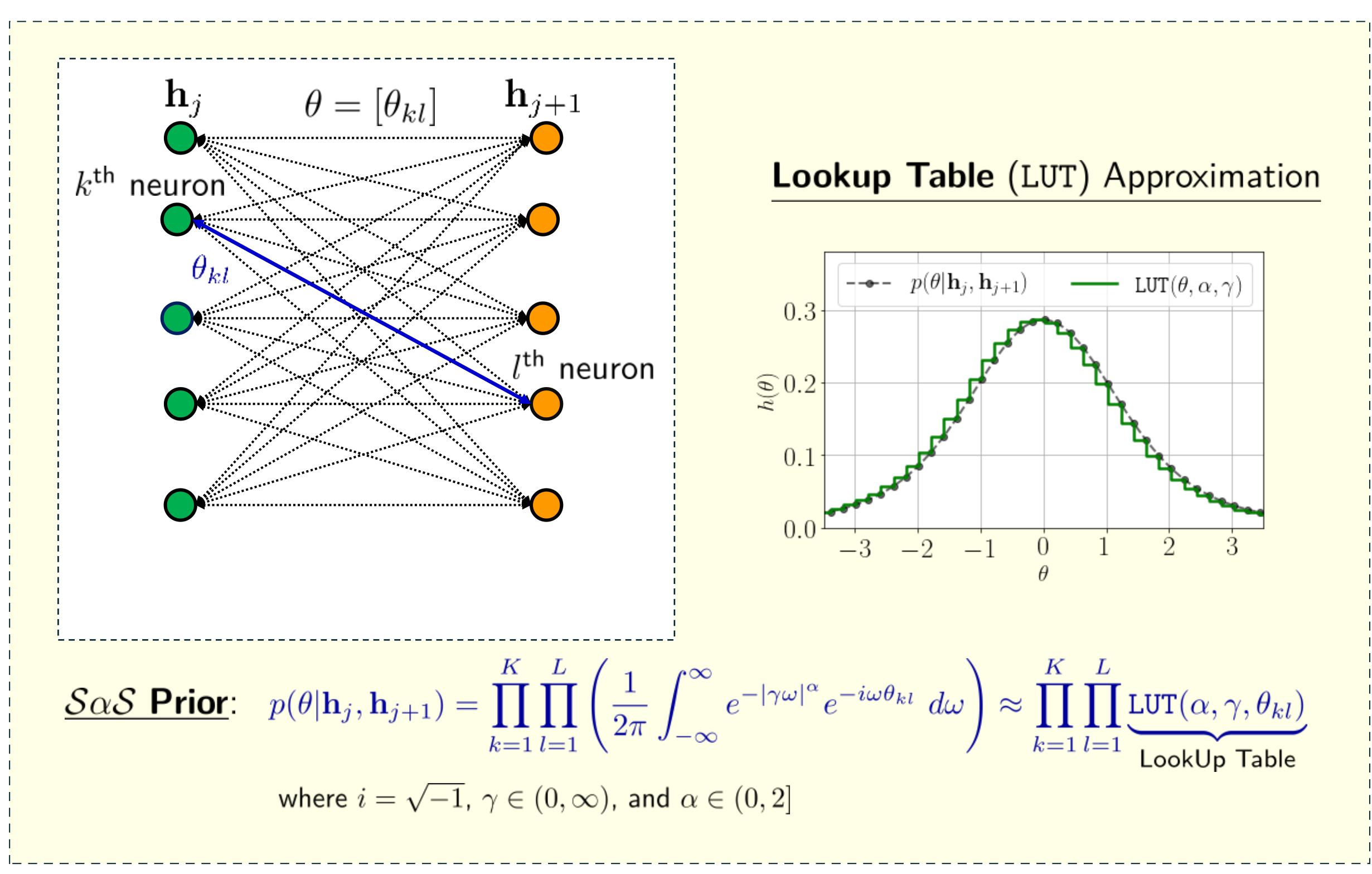

(b) Between-layer $\mathcal{S}\alpha\mathcal{S}$ weight priors with lookup-table approximation

Figure 1 : Between-layer $\mathcal{S}\alpha\mathcal{S}$ weight prior with stability $\alpha = 0.5$ and dispersion $\gamma = 1.0$: The soft diamond shape of $\mathcal{S}\alpha\mathcal{S}$ priors stems from the weight constraint sets on the corresponding log-priors. (a) shows the thick-tailed sub-Cauchy soft-diamond for $\alpha = 0.5$ that tended to perform best on image and text-translation test sets and that produced sparse sets of trained synaptic weights. (b) shows how a lookup table can approximate a $\mathcal{S}\alpha\mathcal{S}$ prior with no known closed form (although all $\mathcal{S}\alpha\mathcal{S}$ priors do have a characteristic function or Fourier transform with a simple complex-exponential form).

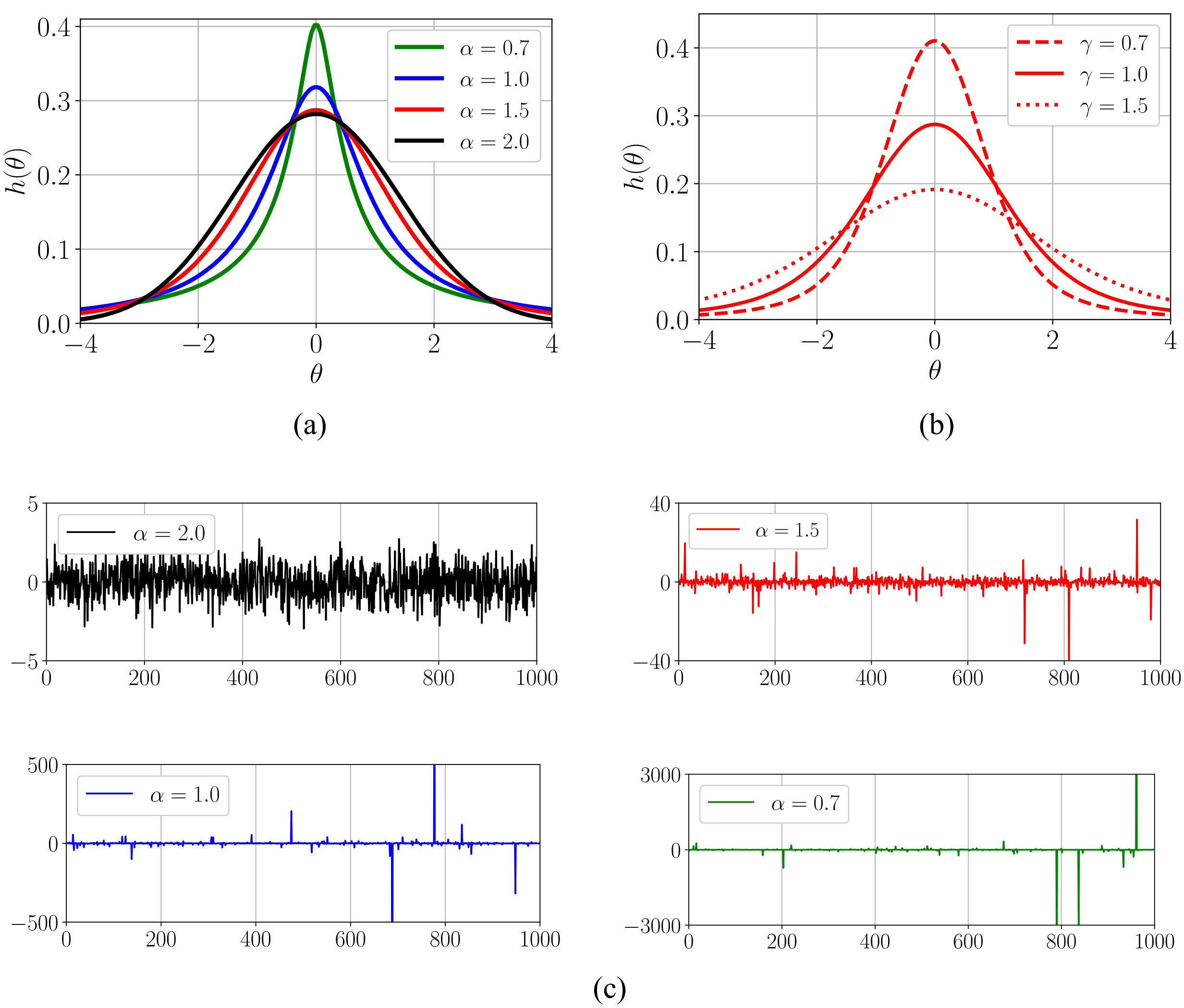

Figure 2: Symmetric alpha-stable $\mathcal{S}\alpha\mathcal{S}$ probability densities $h(\theta)$ with dispersion $\gamma$, stability $\alpha$, and location $\mu = 0$: (a) shows the 4 bell-curve densities for $\gamma = 1.0$ and $\alpha \in \{0.7, 1.0, 1.5, 2.0\}$, (b) shows the 3 densities for $\alpha = 1.5$ and $\gamma \in \{0.7, 1.0, 1.5\}$, and (c) shows 1,000 white-noise samples from each $\mathcal{S}\alpha\mathcal{S}$ density with $\alpha \in \{0.7, 1.0, 1.5, 2.0\}$, $\mu = 0$, and $\gamma = 1.0$. The noise becomes much more impulsive as $\alpha$ falls.

propagates the approximation error from the output layer back through the hidden layers to the input layer [5]–[7].

Bayesian BP generalizes BP training so that it maximizes the network posterior $p(y|x, \theta)$ given prior constraints on synaptic weights and perhaps other parameters. It penalizes or regularizes BP training and tends to give better performance but at slightly higher computational cost. Bayesian BP adds a scaled log-prior or penalty term to the log-likelihood:

$$\hat{\theta} = \arg\max_{\theta} \; \ln p(y|x,\theta) \; + \; c \ln p(\theta|x) \tag{2}$$

where $p(\theta|x)$ denotes the weight prior and $c > 0$ is the log-prior rate. The Bayesian BP framework further extends to the bidirectional case using the same synaptic web but with forward and backward flow of neural signals. Bidirectional BP minimizes the directional errors for the forward and backward signal flow over a deep network [8]. Bayesian bidirectional BP adds a scaled log-prior to the sum of forward and backward loglikelihoods [9].

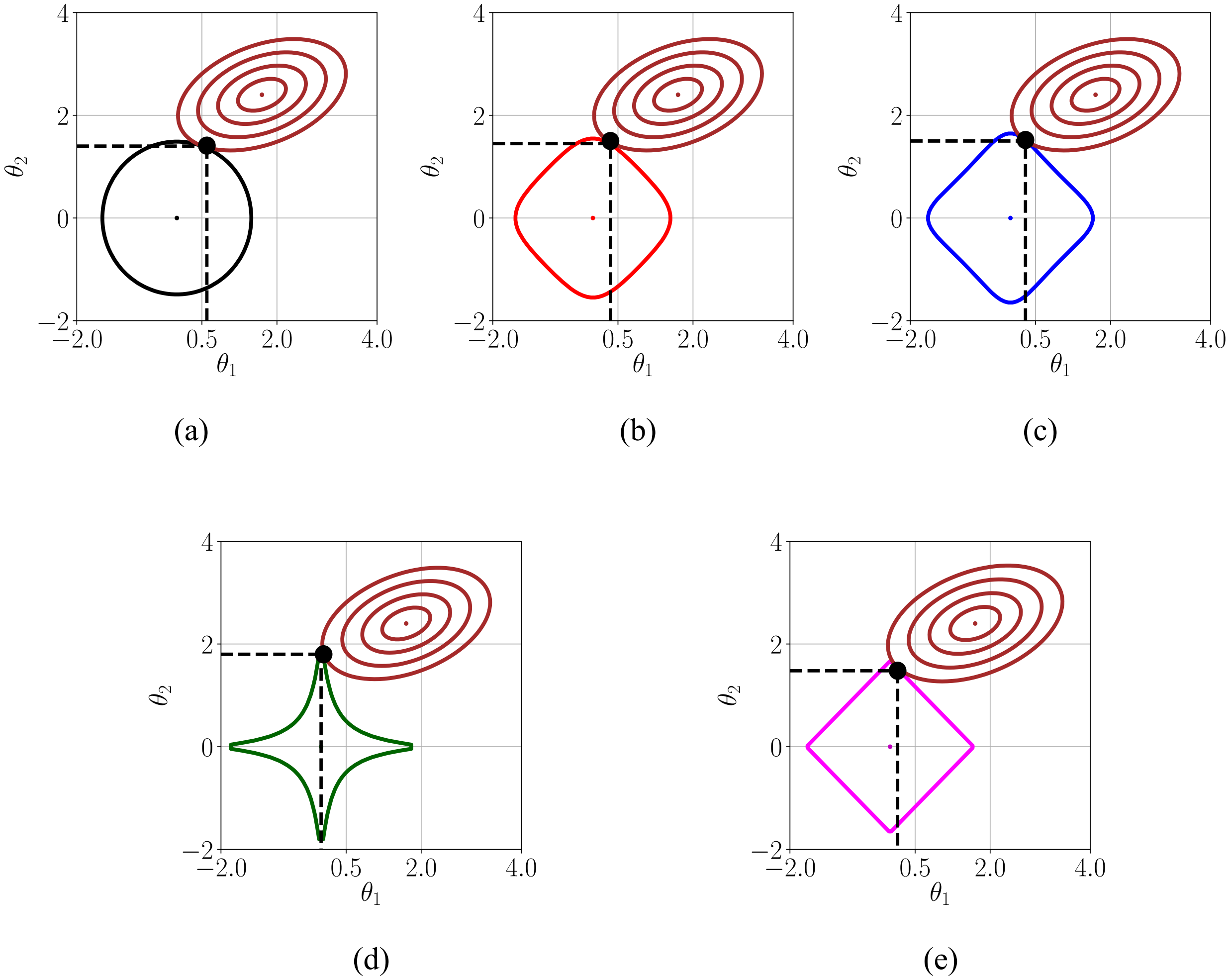

Figure 3: Sparsity with soft-diamond $\mathcal{S}\alpha\mathcal{S}$ weight priors: These plots show the solutions of a least-squared-error model with $\mathcal{S}\alpha\mathcal{S}$ priors as constraints. The weight sparsity grows as the bell-curve tail-thickness value $\alpha$ value falls. (a) shows the solution with Gaussian value $\alpha$ = 2.0. The constraint shape is a ball and the solution is $\theta_1 = 0.6$ and $\theta_2 = 1.41$, (b) shows the solution with $\alpha = 1.5$. The constraint shape is a soft diamond and the solution is $\theta_1 = 0.35$ and $\theta_2 = 1.5$, (c) shows the solution with Cauchy value $\alpha = 1.0$. The constraint shape is a much softer diamond and the solution is $\theta_1 = 0.3$ and $\theta_2 = 1.52$, (d) shows the solution with sub-Cauchy value $\alpha = 0.5$. The constraint shape is soft star and the solution is $\theta_1 = 0.02$ and $\theta_2 = 1.8$, and (e) shows the solution with a non- $\mathcal{S}\alpha\mathcal{S}$ Laplacian prior. The constraint shape is a hard diamond and the solution is $\theta_1 = 0.15$ and $\theta_2 = 1.48$.

Adding weight priors to Bayesian BP tends to improve the performance of deep neural network models. The benefits include better generalization [10], [11], stable and robust training [12], [13], weight sparsity [14]–[16], and faster training. Weight priors also improve the performance of deep networks with post-training weight pruning. Han *et. al* [17] found that using $\ell_1$ or $\ell_2$ regularization improves the performance of deep neural classification with post-training weight pruning.

Gaussian priors ($\ell_2$ regularizers) and Laplacian priors ($\ell_1$ regularizers) remain widely used parametric weight priors in BP training. We here explore $\mathcal{S}\alpha\mathcal{S}$ bell-curve parametric priors as alternatives to Gaussian and Laplacian priors.

We first review the underlying thick-tailed symmetric alpha-stable bell curves [18]. The characteristic function for an alpha stable distribution with stability $\alpha \in (0,2]$, symmetry $\beta \in [-1,1]$, location $\mu \in \mathbb{R}$, and dispersion $\gamma \in \mathbb{R}^+$ is

$$\varphi(\omega; \alpha, \beta, \mu, \gamma) = e^{-i\omega\mu - |\gamma\omega|^{\alpha}(1 - i\beta\,\mathrm{sgn}(\omega)\Phi)} \tag{3}$$

where

$$\Phi = \begin{cases} \tan\left(\frac{\pi\alpha}{2}\right), & \alpha \neq 1 \\ -\frac{2}{\pi}\log|\omega|, & \alpha = 1 \end{cases} . \tag{4}$$

The only two symmetric closed-form bell curves in this family are the Gaussian ($\alpha = 2$) with thin exponential tails and the Cauchy ($\alpha = 1$) with thick power-law tails. Tail thickness increases as the parameter $\alpha$ falls from 2 to just above 0. Figure 2(a) shows four such bell curves. Figure 2(c) shows the corresponding white-noise plots. The noise fluctuations increase substantially as the parameter $\alpha$ falls. Figure 2(b) shows three $\alpha = 1.5$ bell curves with different widths or dispersions. All stable bell curves have a finite dispersion but only the Gaussian has finite variance.

Most $\mathcal{S}\alpha\mathcal{S}$ densities do not have a closed form. There are exceptions [19]–[22] as in the above symmetric Gaussian and Cauchy densities and the asymmetric Levy stable density. This lack of a closed-form probability density functions for $\mathcal{S}\alpha\mathcal{S}$ priors make it hard to compute the derivative of log-priors. We found that a simple lookup table overcame this problem and made training with soft diamond regularizers practical.

We tested the new soft-diamond regularizers on classification accuracy and post-pruning behavior. Weight pruning favored the $\alpha = 1.5$ soft diamond only the simplest image cast of CIFAR-10. The Cauchy and $\alpha = 0.5$ soft diamonds performed best for high sparsity and accuracy without post training pruning. The sub-Cauchy $\alpha = 0.5$ soft diamond performed best again on both image and language translation tasks and metrics. The last sections below present the findings for translating German text to English text.

### Section II: Bayesian Backpropagation with $\mathcal{S}\alpha\mathcal{S}$ Priors

Bayesian BP training with soft diamonds adds a scaled log-prior to the log-likelihood as in (2) but uses a prior $p(\theta|x)$ that is $\mathcal{S}\alpha\mathcal{S}$. Training maximizes the sum of the log-likelihood and scaled log-prior using stochastic gradient ascent:

$$\theta^{(t+1)} = \theta^{(t)} + \lambda_t(\nabla_\theta \ln p(y|x,\theta) + c\,\nabla_\theta \ln p(\theta|x)) \tag{5}$$

at $\theta = \theta^{(t)}$ if $\theta^{(t)}$ and $\lambda_t$ are the respective weights and learning rate after $t$ training iterations for $c > 0$ and where $p'(\theta|x) = \nabla_\theta p(\theta|x)$.

The stable probability density $h(\theta)$ equals the Fourier transform of its characteristic function $\varphi(\omega;\alpha,\beta,\mu,\gamma)$ in (3):

$$h(\theta) = \frac{1}{2\pi}\int_{-\infty}^{\infty} \varphi(\omega;\alpha,\beta,\mu,\gamma)\, e^{-i\omega\theta}\, d\omega. \tag{6}$$

The symmetry $\beta$ equals 0 for $\mathcal{S}\alpha\mathcal{S}$ densities. So

$$\varphi(\omega;\alpha,\beta=0,\mu,\gamma) = e^{i\omega\mu-|\gamma\omega|^\alpha} \tag{7}$$

and this gives

$$h(\theta) = \frac{1}{2\pi}\int_{-\infty}^{\infty} e^{i\omega\mu-|\gamma\omega|^\alpha}\; e^{-i\omega\theta}\, d\omega. \tag{8}$$

Figure 3 shows the shape of the weight constraint set for $\mathcal{S}\alpha\mathcal{S}$ priors with squared-error optimization. The priors that give a soft-diamond shape promote sparsity because the sharper diamond "points" with a smaller $\alpha$ favor zeroing-out weight parameters. Figure 4 compares the geometry of the weight constraint sets for different $\mathcal{S}\alpha\mathcal{S}$ priors. The constraints follow from $\ln h(\theta_1) + \ln h(\theta_2) = \kappa$ where $\kappa$ is a constant and $h$ is the corresponding prior. Figure 5 shows how the value of $\gamma$ affects the geometry of the weight constraint. This chapter uses $\mathcal{S}\alpha\mathcal{S}$ priors with $\mu = 0$.

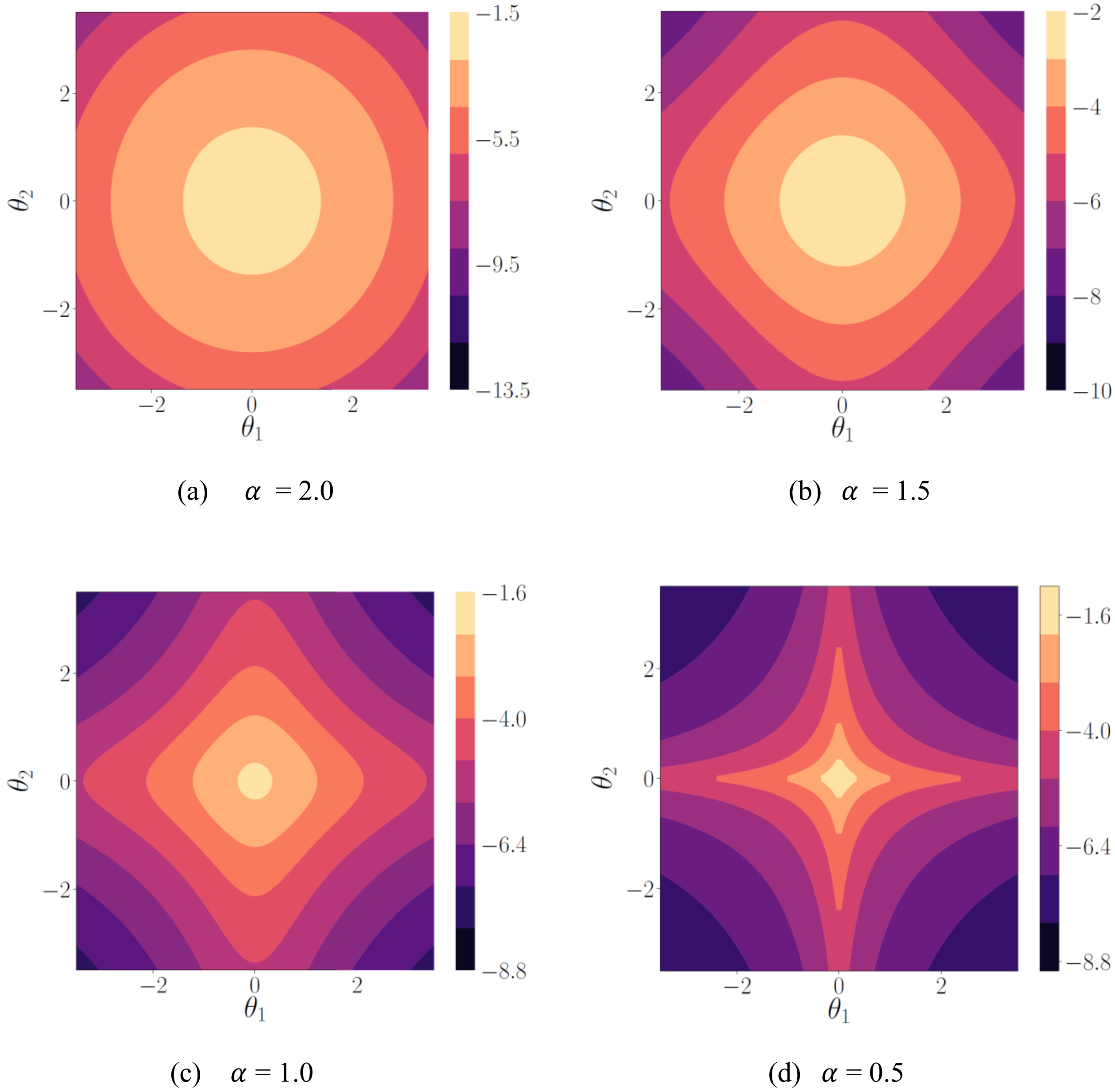

(a) $\alpha = 2.0$ (b) $\alpha = 1.5$

(c) $\alpha = 1.0$ (d) $\alpha = 0.5$

Figure 4: Shapes of $\mathcal{S}\alpha\mathcal{S}$ weight constraint sets for dispersion value $\gamma = 1$ with location $\mu = 0$: A smaller $\alpha$ value or bell-curve thickness gives sharper diamonds. The shape evolves from a ball in the Gaussian case $\alpha = 2$ to a softer diamond in the Cauchy case $\alpha = 1$ and on down to a star in the sub-Cauchy case with $\alpha = 0.5$.

The next section addresses the problem that the derivative of most $\mathcal{S}\alpha\mathcal{S}$ densities also lack any known closed form.

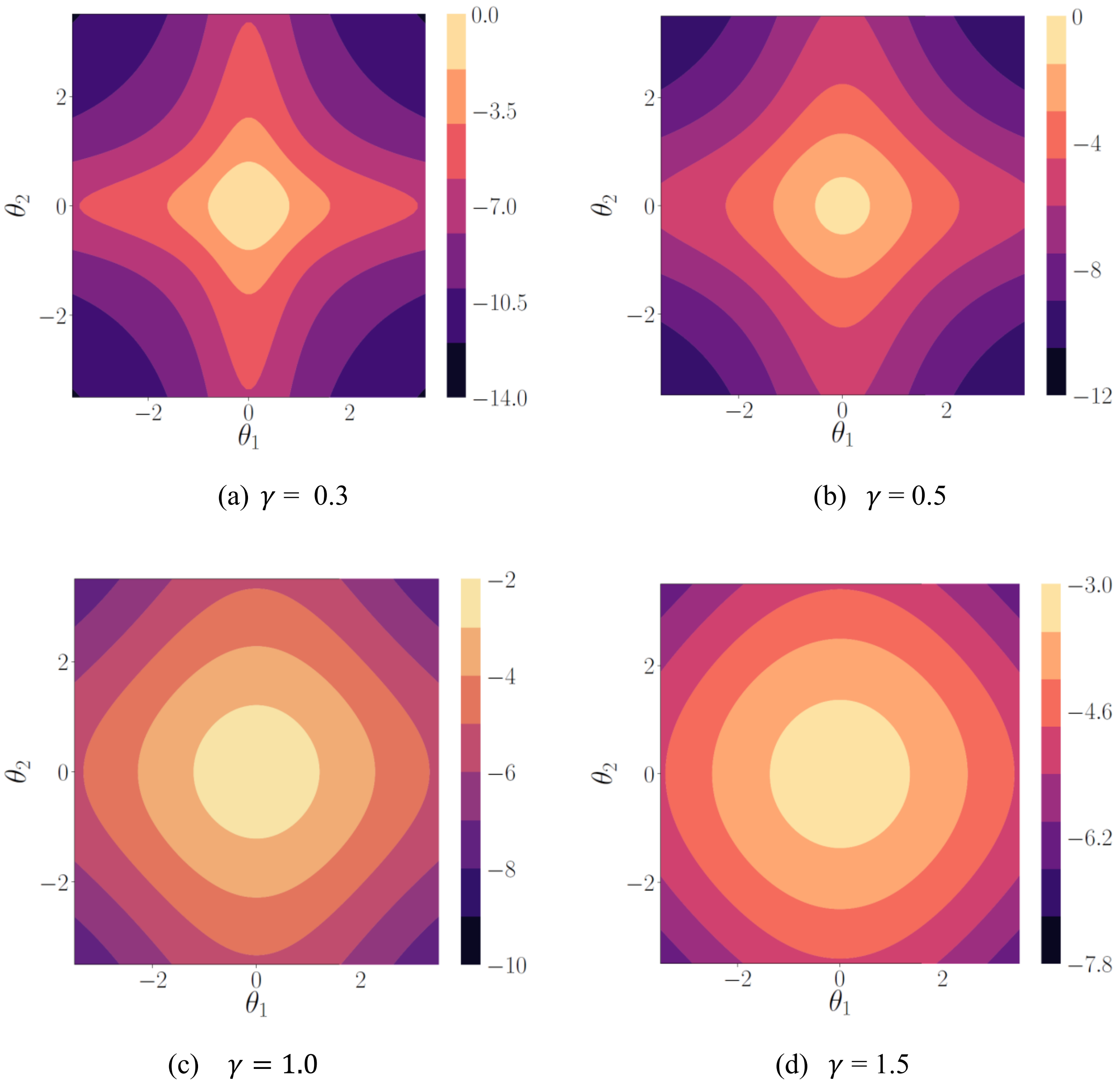

(a) $\gamma = 0.3$ (b) $\gamma = 0.5$

(c) $\gamma = 1.0$ (d) $\gamma = 1.5$

Figure 5: How dispersion values $\gamma$ affect the shape of $\mathcal{S}\alpha\mathcal{S}$ weight constraint sets for bell curves with $\alpha = 1.5$ and location $\mu = 0$: The shapes evolve from a star with $\gamma = 0.3$ to a soft diamond with $\gamma = 1.0$ and to a rounded square with $\gamma = 1.5$.

### Section III: Approximating the Derivative of the Log of $\mathcal{S}\alpha\mathcal{S}$ Priors

This section presents an efficient way to compute the log-prior derivative of $\mathcal{S}\alpha\mathcal{S}$ densities using the finite difference method. The approach defines a lookup table for the values of the derivative of the log-priors. The central finite difference estimates the derivative over a set of fixed points. These values approximate the derivative over the entire domain of the parameter space.

Define $f$ as a real-valued function on closed interval $[a, b]$. Then the quotient for any $x \in [a,b]$ is

$$\phi(t) = \frac{f(t) - f(x)}{t - x} \tag{9}$$

where $a < t < b$ and where $t \neq x$ is such that

$$f'(x) = \lim_{t \to x} \frac{f(t) - f(x)}{t - x} = \lim_{t \to x} \phi(t) . \tag{10}$$

Put $g(\theta) = \ln p(\theta|x)$ for $g(\theta)$. Then use the central finite difference method to approximate the derivative of $g$ at $\theta$ [23]:

$$g'(\theta) = \frac{\mathrm{d}\, g(\theta)}{\mathrm{d}\theta} \approx \frac{g(\theta + \delta) - g(\theta - \delta)}{2\delta} . \tag{11}$$

The approximation error is $O(\delta^2)$ [24], [25]. We define a bounded region $[-\epsilon, \epsilon]$ and divide it into $2N_{\mathrm{g}}$ steps where $\epsilon \in \mathbb{R}^+$ and $N_{\mathrm{g}} \in \mathbb{Z}^+$. So the step size $\delta$ is

$$\delta = \frac{\epsilon}{N_{\mathrm{g}}} . \tag{12}$$

Figure 6 shows these steps over the interval $[-1,1]$ for the standard Cauchy $\mathcal{S}\alpha\mathcal{S}$ density with $\alpha = 1.0$ and $\sigma = 1.0$.

The key $\mathcal{T}_K$ for $\theta \in \mathbb{R}$ takes on a value from $\{-N_{\mathrm{g}}, .., 0, .., N_{\mathrm{g}}\}$:

$$\mathcal{T}_K(\theta) = \begin{cases} -N_{\mathrm{g}}, & -\infty < \theta \leq -N_{\mathrm{g}} \\ \left\lfloor \dfrac{\theta}{\epsilon} \right\rceil, & -N_{\mathrm{g}} < \theta \leq N_{\mathrm{g}} \\ N_{\mathrm{g}}, & N_{\mathrm{g}} < \theta < \infty \end{cases} \tag{13}$$

yields keys restricted to the small finite interval $[-\epsilon, \epsilon]$ instead of to the whole domain $(-\infty, \infty)$ of an $\mathcal{S}\alpha\mathcal{S}$ density.

Figure 7 shows the synaptic weight distribution for deep neural classifiers before training and after training. The weights use the Xavier uniform initialization method [26]. The graphs show

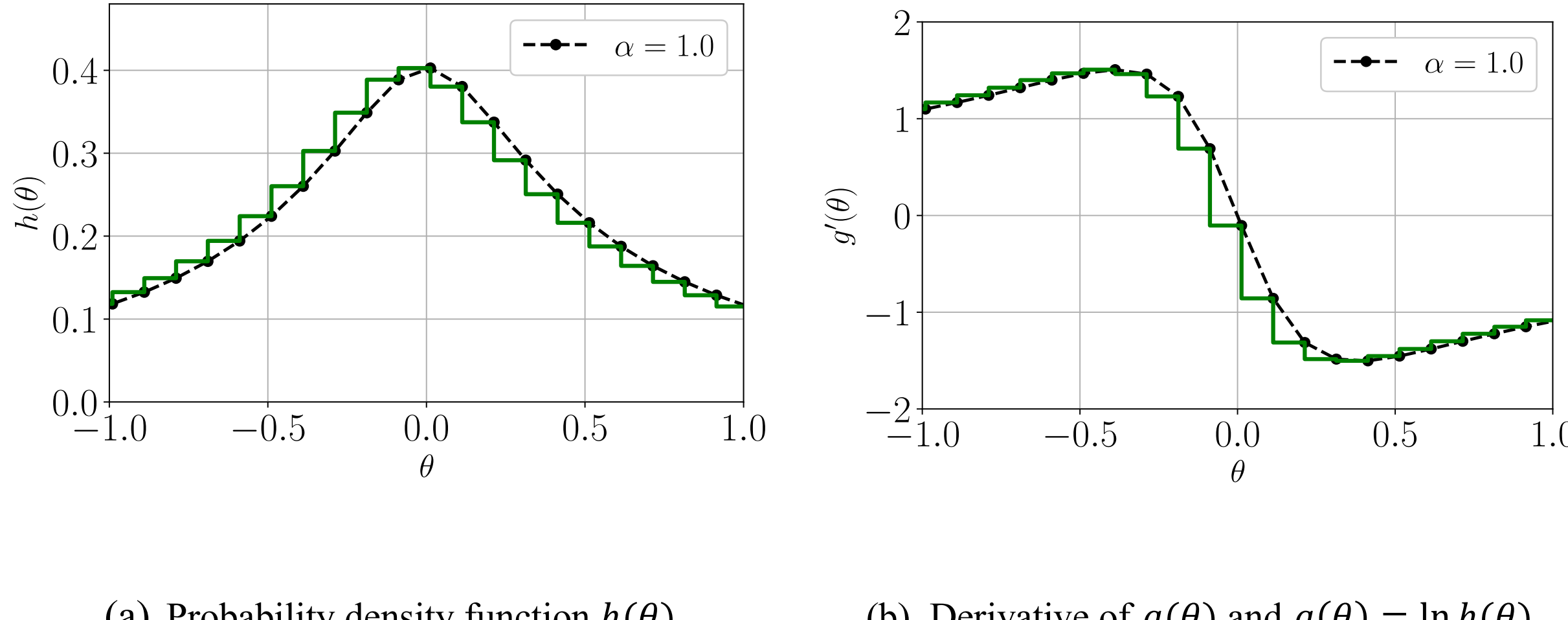

(a) Probability density function $h(\theta)$ (b) Derivative of $g(\theta)$ and $g(\theta) = \ln h(\theta)$

Figure 6: Step size for derivative lookup table with quantized interval $[-1,1]$ for the standard Cauchy $\mathcal{S}\alpha\mathcal{S}$ density with $\alpha = 1.0$ and $\gamma = 1.0$. The number of steps is $2N_g = 20$ with step size is $\delta = 0.1$. (a) shows the quantized Cauchy density and (b) shows the approximated derivative of the Cauchy log-prior.

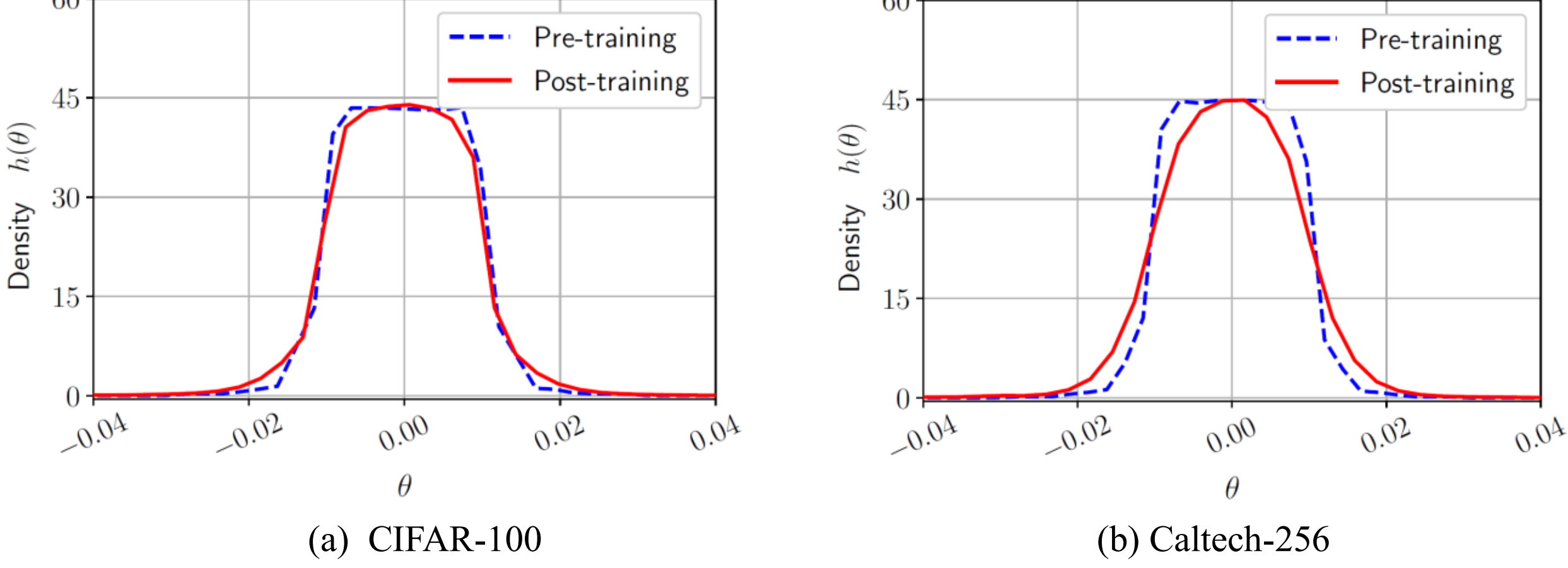

(a) CIFAR-100 (b) Caltech-256

Figure 7: Weight distribution with uniform prior for deep neural classifiers before training and after training over 50 training epochs. The weight values fell between the values $-0.03$ and $0.03$. (a) shows the weight distribution before and after training on CIFAR-100 dataset. (b) shows the weight distribution before and after training on Caltech-256 dataset.

that the interval $[-0.03, 0.03]$ bounds the synaptic weights in these two cases. The small finite bound for $\theta$ suggests that we need only a small $\epsilon$ value to cover the synaptic weight domain.

The corresponding value $\mathcal{T}_V(\mathcal{T}_K(\theta))$ for key $\mathcal{T}_K(\theta)$ assigns an approximate derivative value to the key. Then the central finite difference method gives

$$\mathcal{T}_V\big(\mathcal{T}_K(\theta)\big) = \frac{p(\mathcal{T}_K(\theta) + \delta|x) - p(\mathcal{T}_K(\theta) - \delta|x)}{2\delta\, p(\mathcal{T}_K(\theta)|x)} \tag{14}$$

where $\mathcal{T}_K(\theta)$ is as in (13). The estimate is a function of its $\mathcal{S}\alpha\mathcal{S}$ density $p(\theta|x)$.

Using a lookup table reduced the computation to estimate derivative of the $\mathcal{S}\alpha\mathcal{S}$ log-priors. We needed to compute $p(\theta|x)$ only over a fixed set of points with the lookup table. $N_{\mathrm{g}}$ could be as small as 100.

Algorithm 1 shows the pseudocode for training a deep network with BP and a $\mathcal{S}\alpha\mathcal{S}$ weight prior. It combines BP training with the lookup table to approximate the derivative of the synaptic-weight $\mathcal{S}\alpha\mathcal{S}$ log-priors.

## Section IV: Simulations

This section explains the experimental setup that combined tasks, datasets, model architectures, and training methods.

### *Subsection 4.1: Tasks*

We trained deep neural classifiers on image datasets. The classifiers mapped an input image to one of *K* possible target vectors with *K* classes. We observed the effect of $\mathcal{S}\alpha\mathcal{S}$ weight priors on the classification accuracy of the neural classifiers. We also observed the sparsity of the weights after training. We also observed the effect of unstructured weight pruning after training deep neural classifiers with $\mathcal{S}\alpha\mathcal{S}$ priors.

We trained deep neural machine translation (NMT) models for text-to-text translation from German to English. Each network used Long Short-Term Memory (LSTM) units as building blocks. These models mapped German texts to their corresponding English translation. The underlying task of an NMT model is classification. We observed the effect of $\mathcal{S}\alpha\mathcal{S}$ weight priors on the models using the following metrics: Bilingual Evaluation Understudy (BLEU) score [27]–[29], the Metric for Evaluation of Translation with Explicit Ordering (METEOR) score [30], and the Recall-Oriented Understudy for Gisting Evaluation (ROGUE) [31]–[33] scores. We also observed the post-training weight sparsity of the models.

### *Subsection 4.2: Datasets*

We used three image classification datasets: CIFAR-10 [34], CIFAR-100, and Caltech-256 [35]. CIFAR-10 dataset consists of 60,000 color images from 10 categories. It has 10 pattern categories ($K$ = 10): airplane, automobile, bird, cat, deer, dog, frog, horse, ship, and truck. This dataset is

**Algorithm 1**: BP Training with a $\mathcal{S}\alpha\mathcal{S}$ prior using SGD optimizer with momentum

---

**Input:** Data $\mathcal{D} = \{x^{(i)}, y^{(i)}\}$, symmetry $\alpha \in (0, 2]$, dispersion $\sigma \in \mathbb{R}^+$, prior coefficient $c \in \mathbb{R}^+$, prior bound $b \in \mathbb{R}^+$, grid count $N_g \in \mathbb{Z}^+$, step size $\delta \in \mathbb{R}^+$ , momentum $m \in (0, 1]$, dampening $\tau \in [0, 1)$, and number of training iterations $N_T$

**Output:** Weight $\theta$

*Initialization* : Weight $\theta^{(0)}$ and learning rate $\lambda_0$

1: Define the log-prior lookup table $\mathcal{T} = \{\mathcal{T}_K, \mathcal{T}_V\}$
Use equations (13) and (14)

2: **for** $t = 0$ to $N_T - 1$ **do**

3: Forward pass $N_\theta$: $a^y = N_\theta(x)\Big|_{\theta=\theta^{(t)}}$

4: Update the weights:

$$g_t = \frac{d \ln p(y|x, \theta)}{d\theta}\Bigg|_{\theta=\theta^{(t)}} + c\mathcal{T}_V(\mathcal{T}_K(\theta^{(t)}))$$

5: **if** $(t = 0)$ **then**

6: $\theta^{(t+1)} = \theta^{(t)} + \lambda_t \; g_t$; $\beta_{t+1} = g_t$

7: **else**

8: $\beta_{t+1} = m \; \beta_t + (1 - \tau)g_t$; $\theta^{(t+1)} = \theta^{(t)} + \; \lambda_t\beta_{t+1}$

9: **end if**

10: Update the learning rate

11: **end for**

12: **return** $\theta^{(N_T)}$

---

balanced with 6,000 images per class: 5,000 training samples and 1,000 testing samples. Each image has size 32 × 32 × 3.

CIFAR-100 is also a dataset of 60,000 color images with image size 32×32×3. But the images come from 100 pattern classes ($K$ = 100) with 600 images per class. Each class consists of 500 training images and 100 test images.

The Caltech-256 dataset has 30,607 images from 256 pattern classes ($K$ = 256) with image size 100×100×3. The number of samples varies between 31 and 80 images. The 256 classes consisted of the two super classes *animate* and *inanimate*. The animate superclass contained 69 pattern classes. The inanimate superclass contained 187 pattern classes. We removed the *cluttered* images and reduced the size of the dataset to 29,780 images. We divided the dataset into 23,824 training images and 5,956 test images. The images had different dimensions. We resized each image to 100×100×3.

We used image augmentation techniques on the training images. These techniques include image flip, image cutout, and channel normalization.

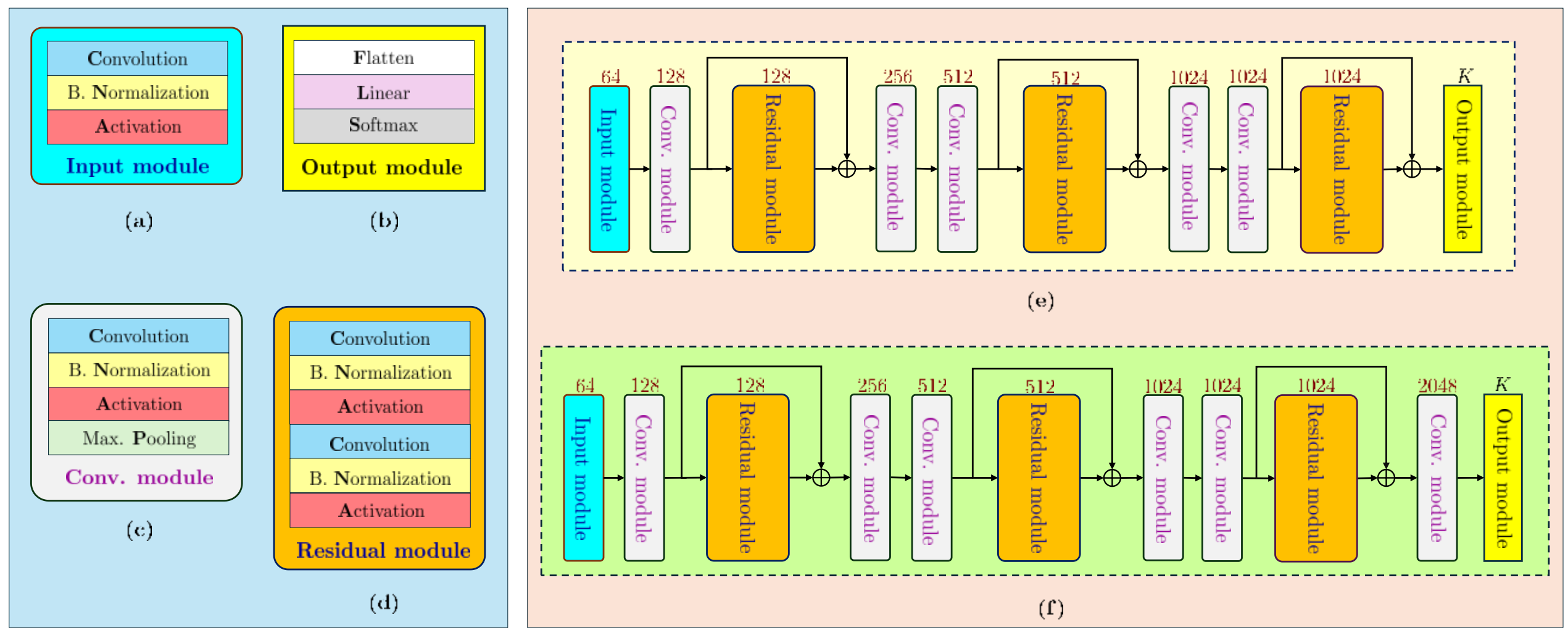

Figure 8: Architecture of the tested deep neural classifiers. Each module's number corresponds to the number of output channels where $K$ was number of class patterns. (a) is the input module. (b) is the output module. (c) is the convolution module. (d) is the residual module. (e) is the full model architecture of the neural classifiers that trained on the CIFAR-10 and CIFAR-100 datasets. (f) is the full model architecture of the neural classifiers that trained on the Caltech-256 dataset.

We used the IWSLT-2016 Evaluation Campaign Dataset [36] for NMT tasks. This dataset includes parallel text translations for multiple languages. We considered the German-English pair in this work. The dataset includes 29,000 training pairs and 1,000 testing pairs of German text and their corresponding English translations.

### *Subsection 4.3: Model Architectures and Training*

The neural classifiers each used convolutional and residual network architecture. Figure 8(a)−(d) show the modules for building the deep neural classifiers. The input module, convolution module, residual module, and output module.

The input module takes in the input image and applies 2D convolution, batch normalization, and a nonlinear activation. The convolution module transforms the hidden features. It is similar in structure to the input module and includes a maximum pooling layer. The residual module is a concatenation of two input modules into one and includes a skip connection. We used rectified linear units or ReLUs as hidden activations.

Figure 8(e) shows the architecture of the neural classifiers that trained on CIFAR-10 and CIFAR-100 datasets. Figure 8(f) shows the architecture of the neural classifiers that trained on Caltech-256.

Stochastic gradient descent trained the models with momentum [37]. Each model trained over 50 epochs. We used a piecewise linear learning rate scheduler. We defined the lookup table for training our classifiers with $\delta = 0.002$ and $N_g = 400$. Algorithm 1 details the training method. We trained the neural classifiers on a V100 GPU.

The NMT models used LSTM units. Figure 9 illustrates the architecture of both an LSTM unit and the overall NMT model. Each machine translation model consisted of two subnetworks: an encoder and a decoder. Both encoder and decoder networks contained three LSTM units each. Each LSTM used an embedding size of 300 and hidden size of 512.

We processed the texts using spaCy's pre-trained tokenizers [38]. The encoder input size was 7,353 and the input decoder size was 5,893 as a result of these tokenizers. We set the minimum frequency to 2.

Each model trained using Stochastic gradient descent with momentum [37]. We used this method to add the lookup table for $\mathcal{S}\alpha\mathcal{S}$ priors. Algorithm 1 presents the pseudocode for this method. Each model trained over 200 epochs with batch size of 128, dropout rate of 0.3, and one V100 GPU.

## Section 5: Results and Discussion

This section explains the experimental results and performance of soft diamond regularizers.

### Subsection 5.1: Image Classification

Table 1 compares the performance of $\mathcal{S}\alpha\mathcal{S}$ weight priors on CIFAR-10 classification. $\mathcal{S}\alpha\mathcal{S}$ priors improved the classification performance and $\alpha = 1.0$ performed best out of all the values we considered. The $\ell_2$ regularizer corresponds to using the Gaussian $\mathcal{S}\alpha\mathcal{S}$ prior with $\alpha = 2$. The $\ell_1$ regularizer corresponds to using Laplace prior. Soft diamond regularizers outperformed both the $\ell_1$ and $\ell_2$ regularizers.

*Effect of the tail-thickness parameter $\alpha$ of $\mathcal{S}\alpha\mathcal{S}$ priors*: Table 1 shows how the soft diamond regularizers performed on CIFAR-10, CIFAR-100, and Caltech-256 classification without post-training weight pruning. It shows the benefit of using $\mathcal{S}\alpha\mathcal{S}$ weight priors as it outperforms $\ell_2$ and $\ell_1$. We considered $\alpha \in \{0.3, 0.5, 1.0, 1.5\}$ The best $\alpha$ varied for the datasets. It depended on the interaction with the dispersion $\gamma$ and the log-prior scale $c$. The sub-Cauchy prior $\alpha = 0.5$ outperformed $\ell_2$ and $\ell_1$ on all the datasets. Table 2 shows that soft-diamond priors always improved the classification accuracy compared to the popular regularization techniques: dropout, batch normalization, and data augmentation on the CIFAR-10 dataset.

Figure 10 shows the impact of soft regularizers on unstructured and post-training weight pruning. They outperformed $\ell_2$ regularizers on the three test sets. Figure 11 shows that $\mathcal{S}\alpha\mathcal{S}$ weight-prior promotes sparsity. The degree of sparsity increases as $\alpha$ decreases.

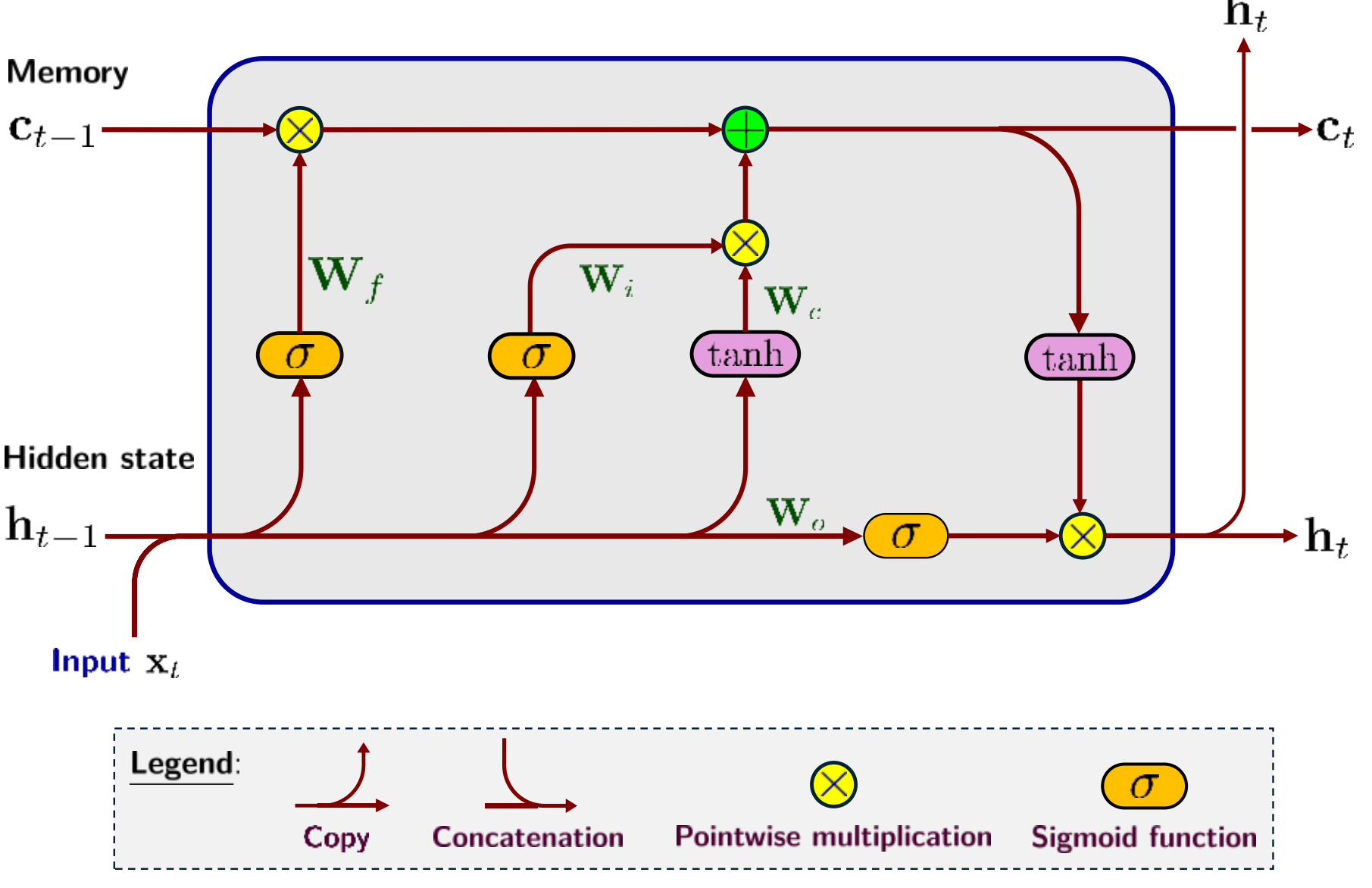

(a) Long Short-Term Memory (LSTM) Unit

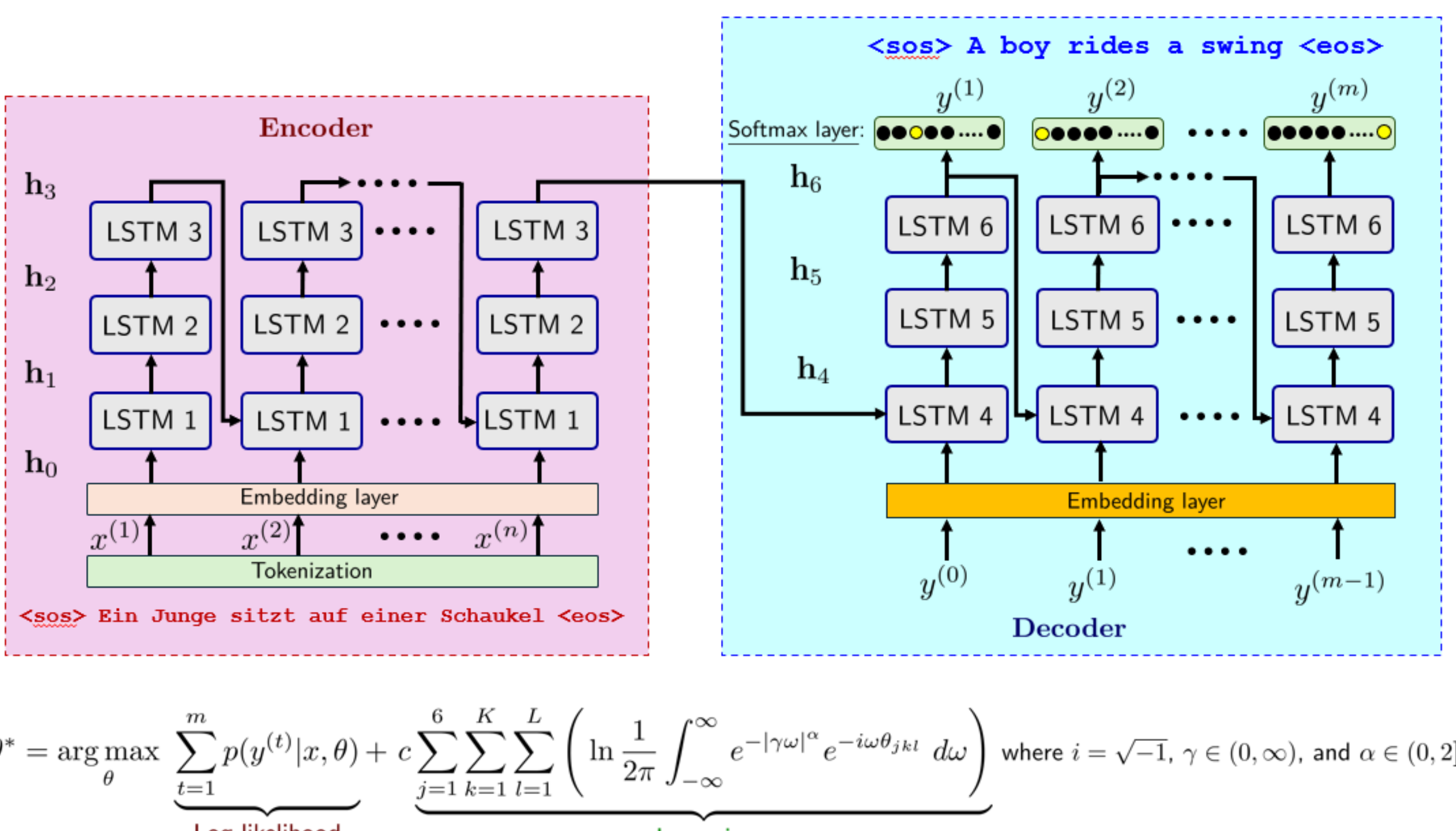

(b) Encoder-decoder network for natural language translation

Figure.9: LSTM network with $S\alpha S$ weight prior

Table 1: Image classification with deep convolutional neural classifiers without post-training weight pruning: We used $\mathcal{S}\alpha\mathcal{S}$ weight priors with location $\mu = 0$, stability $\alpha$, and dispersion $\gamma$. The baseline models used uniform weight prior (no regularizer) and their respective classification accuracy was 91.31% for CIFAR-10, 72.74% for CIFAR-100, and 61.88% for Caltech-256. Each classifier trained over 50 epochs with the stochastic gradient descent optimizer using Algorithm 1.

| **Dataset** | **Prior** | **Classification Accuracy** | | | | | |
|---|---|---|---|---|---|---|---|
| | | $\boldsymbol{\gamma = 0.1}$ | $\boldsymbol{\gamma = 0.3}$ | $\boldsymbol{\gamma = 0.5}$ | $\boldsymbol{\gamma = 1.0}$ | $\boldsymbol{\gamma = 1.5}$ | $\boldsymbol{\gamma = 2.0}$ |
| **CIFAR-10** | Gaussian | 94.64% | 94.51% | **95.11%** | 95.08% | 94.81% | 94.50% |
| | $\mathcal{S}\alpha\mathcal{S}$: $\alpha = 1.5$ | 94.60% | 94.97% | **95.88%** | 94.88% | 94.89% | 94.81% |
| | $\mathcal{S}\alpha\mathcal{S}$: $\alpha = 1.0$ | 94.33% | 94.55% | 94.52% | 94.83% | 94.97% | **94.99%** |
| | $\mathcal{S}\alpha\mathcal{S}$: $\alpha = 0.5$ | 94.66% | 94.74% | 94.75% | 94.68% | **94.76%** | 94.56% |
| | $\mathcal{S}\alpha\mathcal{S}$: $\alpha = 0.3$ | 94.72% | 94.83% | 94.71% | **94.90%** | 94.72% | 94.61% |
| | Laplacian | 93.82% | 93.78% | 93.64% | **93.84%** | 94.72% | 94.61% |
| **CIFAR-100** | Gaussian | 74.63% | 74.39% | 75.50% | 75.40% | 75.62% | 75.48% |
| | $\mathcal{S}\alpha\mathcal{S}$: $\alpha = 1.5$ | 76.29% | 75.26% | 75.11% | 74.70% | 75.37% | 75.49% |
| | $\mathcal{S}\alpha\mathcal{S}$: $\alpha = 1.0$ | **76.60%** | 76.53% | 75.31% | 75.84% | 75.09% | 75.27% |
| | $\mathcal{S}\alpha\mathcal{S}$: $\alpha = 0.5$ | 76.31% | **76.71%** | **77.01%** | 76.74% | 76.93% | 76.74% |
| | $\mathcal{S}\alpha\mathcal{S}$: $\alpha = 0.3$ | 73.73% | 73.75% | 73.96% | 74.99% | 75.01% | **75.54%** |
| | Laplacian | 76.66% | 76.69% | 76.49% | 76.56% | **76.79%** | 76.67% |
| **Caltech-256** | Gaussian | 66.47% | 66.55% | 66.34% | **66.71%** | 66.08% | 66.50% |
| | $\mathcal{S}\alpha\mathcal{S}$: $\alpha = 1.5$ | 67.26% | 66.47% | 66.08% | 66.87% | 66.77% | **66.92%** |
| | $\mathcal{S}\alpha\mathcal{S}$: $\alpha = 1.0$ | 67.01% | 66.89% | **67.51%** | 66.08% | 66.66% | 66.30% |
| | $\mathcal{S}\alpha\mathcal{S}$: $\alpha = 0.5$ | 66.81% | **68.57%** | 68.55% | 68.28% | 67.18% | 67.08% |
| | Laplacian | **68.20%** | 67.39% | 67.95% | 67.95% | 67.90% | 67.80% |

*Effect of the log-prior scale $c$:* We ran experiments with $c \in (0.0001, 100)$. We considered 15 values over this range. The results show that the best value $c^*$ depends on the values of $\alpha$ and $\gamma$. $c^*$ increases with an increase in dispersion $\gamma$ for a fixed stability $\alpha$. The relationship between $c$ and classification accuracy tends to follow an inverted U-shape for a fixed $\alpha$.

*Effect of the step-size $\delta$ of the SαS lookup table:* We define the step size $\delta$ (Algorithm 1) for training with $\mathcal{S}\alpha\mathcal{S}$ weight priors. Figure 12 shows the effect of $\delta$ on the performance of the deep neural classifier that used the priors. Performance increased as $\delta$ decreased.

## Subsection 5.2: Neural Machine Translation

Table 3 compares the performance of $\mathcal{S}\alpha\mathcal{S}$ weight priors on the German-English translation on the IWSLT-2016 dataset. This table reports the BLEU and METEOR scores on the testing set. $\mathcal{S}\alpha\mathcal{S}$ priors improved the machine translation performance with respect to these metrics and $\alpha = 0.5$ performed best out of all the values we considered.

Table 2: Boosting dropout, batch normalization, and data acquisition regularization in CIFAR-10 classification: $\mathcal{S}\alpha\mathcal{S}$ weight priors always improved the performance of these three regularization techniques. The classifiers trained over 100 epochs. The soft-diamond priors gave the most benefit (93.83% accuracy) when combined with image augmentation.

| **Model Architecture** | **Classification Accuracy** | | | | |
|---|---|---|---|---|---|
| | **Batch sz. = 512** | **Batch sz. = 256** | **Batch sz. = 128** | **Batch sz = 64** | **Batch sz. = 32** |
| Baseline (NN only) | 46.68% | 55.36% | 62.65% | 67.37% | 69.81% |
| SαS | 47.78% | 57.60% | 64.36% | 67.49% | 68.48% |
| Dropout | 61.58% | 69.76% | 69.09% | 72.94% | 75.73% |
| Dropout + SαS | 72.14% | 77.08% | 79.41% | 78.18% | 78.13% |
| Image Aug. | 88.93% | 90.94% | 92.02% | 92.34% | 91.48% |
| Image Aug. + SαS | 90.70% | 92.97% | 93.53% | **93.83%** | 92.36% |
| Batch Norm. | 55.92% | 57.73% | 60.58% | 62.47% | 67.03% |
| Batch Norm. + SαS | 78.95% | 77.56% | 82.17% | 80.96% | 75.61% |

Table 3: BLEU and METEOR performance with LSTM for Text-to-Text translation on the IWSLT 2016 dataset: $\mathcal{S}\alpha\mathcal{S}$ weight priors always improved the performance of the neural translation models. The classifiers trained over 200 epochs.

| **Weight Prior** | **BLEU** | **METEOR** |
|---|---|---|
| Uniform | 0.205 | 0.504 |
| Laplace | 0.233 | 0.510 |
| $\mathcal{S}\alpha\mathcal{S}$: $\alpha = 2.0$ | 0.237 | 0.527 |
| $\mathcal{S}\alpha\mathcal{S}$: $\alpha = 1.5$ | 0.244 | 0.534 |
| $\mathcal{S}\alpha\mathcal{S}$: $\alpha = 1.0$ | 0.249 | 0.538 |
| $\mathcal{S}\alpha\mathcal{S}$: $\alpha = 0.5$ | 0.266 | 0.561 |
| $\mathcal{S}\alpha\mathcal{S}$: $\alpha = 0.3$ | 0.227 | 0.518 |
| $\mathcal{S}\alpha\mathcal{S}$: $\alpha = 0.1$ | 0.211 | 0.508 |

Table 4 compares the effect of $\mathcal{S}\alpha\mathcal{S}$ priors using the ROUGE scores. The table also shows that $\mathcal{S}\alpha\mathcal{S}$ priors improved performance and $\alpha = 0.5$ performed best. Figure 13 shows that $\mathcal{S}\alpha\mathcal{S}$ priors can promote sparsity better than Laplace priors. This compares the post-training distribution of weights. $\mathcal{S}\alpha\mathcal{S}$ prior with $\alpha = 0.5$ forces more weights to take small values than with a Laplace prior. Figures 14-15 shows the effect of the log-prior rate $c$ on the performance of NMT models.

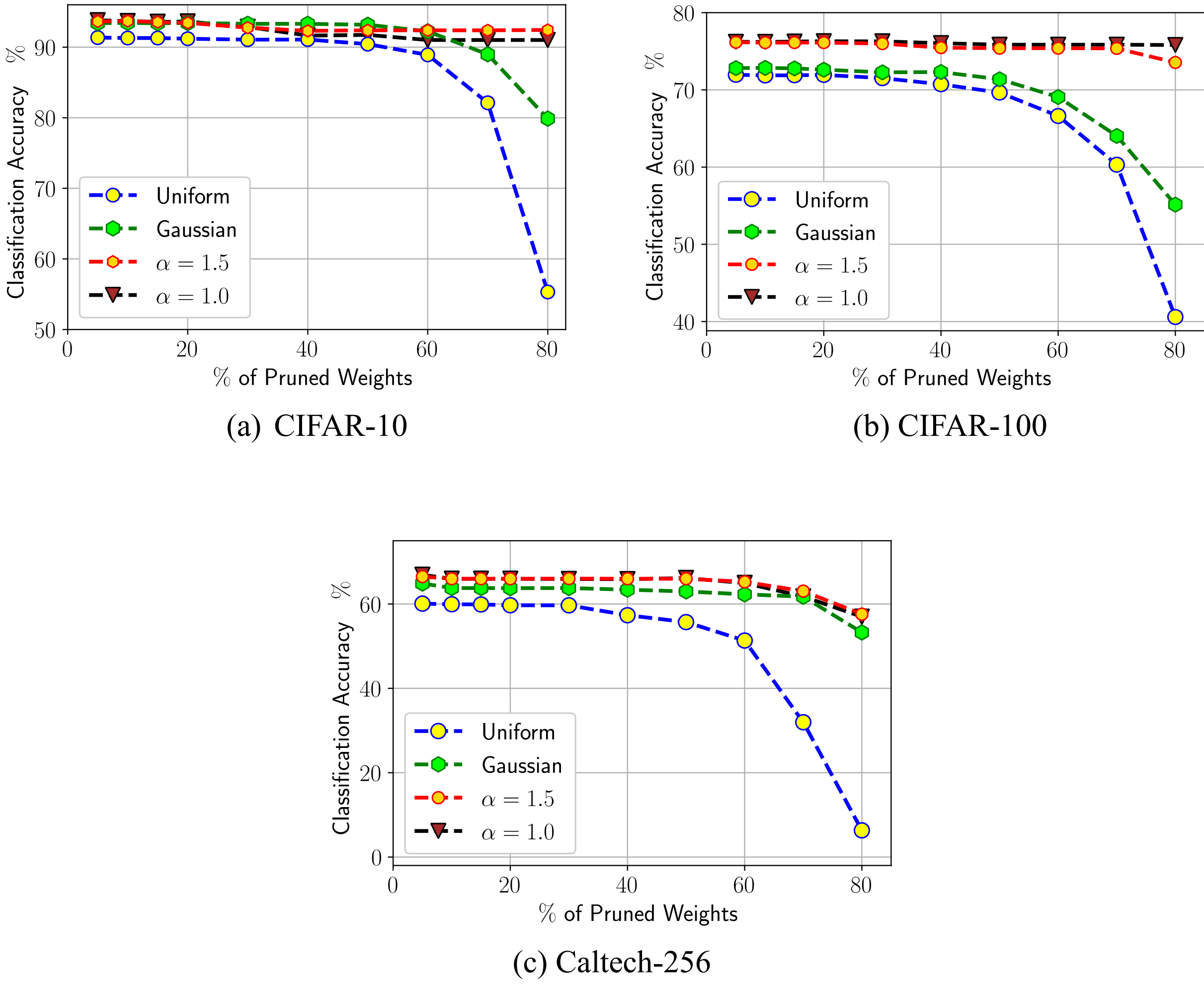

(a) CIFAR-10

(b) CIFAR-100

(c) Caltech-256

Figure 10: Weight pruning effects of $\mathcal{S}\alpha\mathcal{S}$ log-priors on deep neural classifiers after post-training weight pruning: $\mathcal{S}\alpha\mathcal{S}$ priors improved the performance of deep neural classifiers with unstructured and post-training weight pruning.

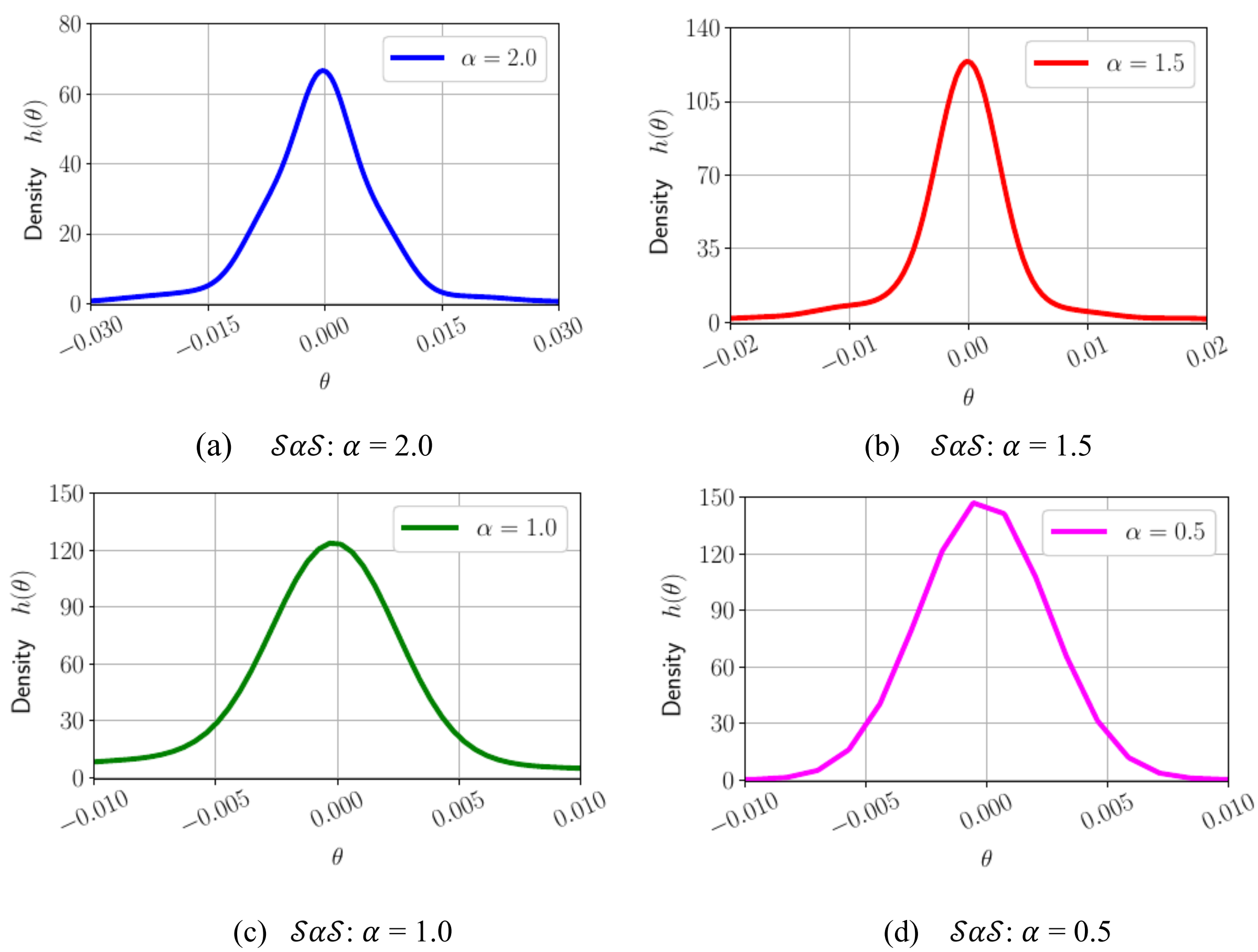

(a) $\mathcal{S}\alpha\mathcal{S}$: $\alpha = 2.0$ (b) $\mathcal{S}\alpha\mathcal{S}$: $\alpha = 1.5$

(c) $\mathcal{S}\alpha\mathcal{S}$: $\alpha = 1.0$ (d) $\mathcal{S}\alpha\mathcal{S}$: $\alpha = 0.5$

Figure 11: Weight distributions of deep neural classifiers after training with $\mathcal{S}\alpha\mathcal{S}$ weight priors using kernel density estimation: The classifiers trained on the CIFAR-100 dataset over 100 training epochs and used a residual network architecture with convolutional layers. Weight sparsity increased as the $\alpha$ of $\mathcal{S}\alpha\mathcal{S}$ tail-thickness parameter $\alpha$ decreased.

Table 4: ROUGE performance with LSTM for Text-to-Text translation on the IWSLT 2016 dataset: $\mathcal{S}\alpha\mathcal{S}$ weight priors always improved the performance of the neural translation models. The classifiers trained over 200 epochs.

| **Weight Prior** | **ROUGE-1** | | | **ROUGE-2** | | | **ROUGE-L** | | |
|---|---|---|---|---|---|---|---|---|---|
| | **Recall** | **Precision** | **F1 score** | **Recall** | **Precision** | **F1 score** | **Recall** | **Precision** | **F1 score** |
| Uniform | 0.556 | 0.516 | 0.531 | 0.292 | 0.285 | 0.287 | 0.539 | 0.496 | 0.511 |
| Laplace | 0.677 | 0.530 | 0.585 | 0.346 | 0.312 | 0.325 | 0.656 | 0.512 | 0.568 |
| $\mathcal{S}\alpha\mathcal{S}$: $\alpha = 2.0$ | 0.637 | 0.537 | 0.576 | 0.332 | 0.312 | 0.319 | 0.631 | 0.469 | 0.529 |
| $\mathcal{S}\alpha\mathcal{S}$: $\alpha = 1.5$ | 0.647 | 0.545 | 0.585 | 0.344 | 0.323 | 0.331 | 0.625 | 0.528 | 0.566 |
| $\mathcal{S}\alpha\mathcal{S}$: $\alpha = 1.0$ | 0.674 | 0.549 | 0.598 | 0.360 | 0.328 | 0.340 | 0.650 | 0.531 | 0.578 |
| $\mathcal{S}\alpha\mathcal{S}$: $\alpha = 0.5$ | 0.681 | 0.568 | 0.613 | 0.369 | 0.341 | 0.352 | 0.655 | 0.547 | 0.590 |
| $\mathcal{S}\alpha\mathcal{S}$: $\alpha = 0.3$ | 0.640 | 0.534 | 0.575 | 0.327 | 0.305 | 0.313 | 0.616 | 0.514 | 0.554 |
| $\mathcal{S}\alpha\mathcal{S}$: $\alpha = 0.1$ | 0.560 | 0.521 | 0.535 | 0.298 | 0.291 | 0.292 | 0.560 | 0.521 | 0.535 |

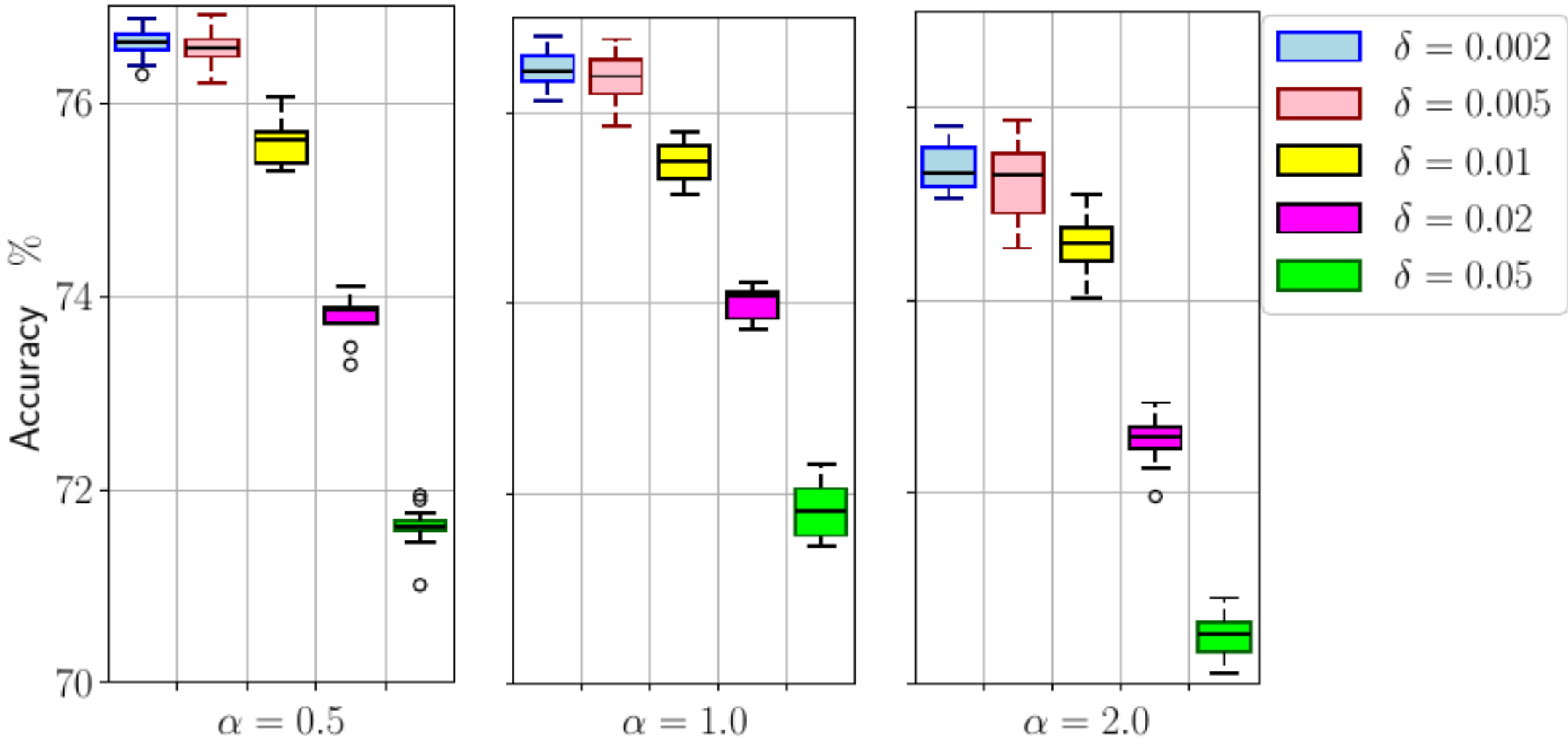

Figure 12: Look-up table step size $\delta$, $\mathcal{S}\alpha\mathcal{S}$ weight priors, and CIFAR-100 classification: The look-up table $\mathcal{T}$ approximates the derivative of the corresponding log-priors. Algorithm 1 combines $\mathcal{T}$ and Bayesian backpropagation to train deep neural classifiers.

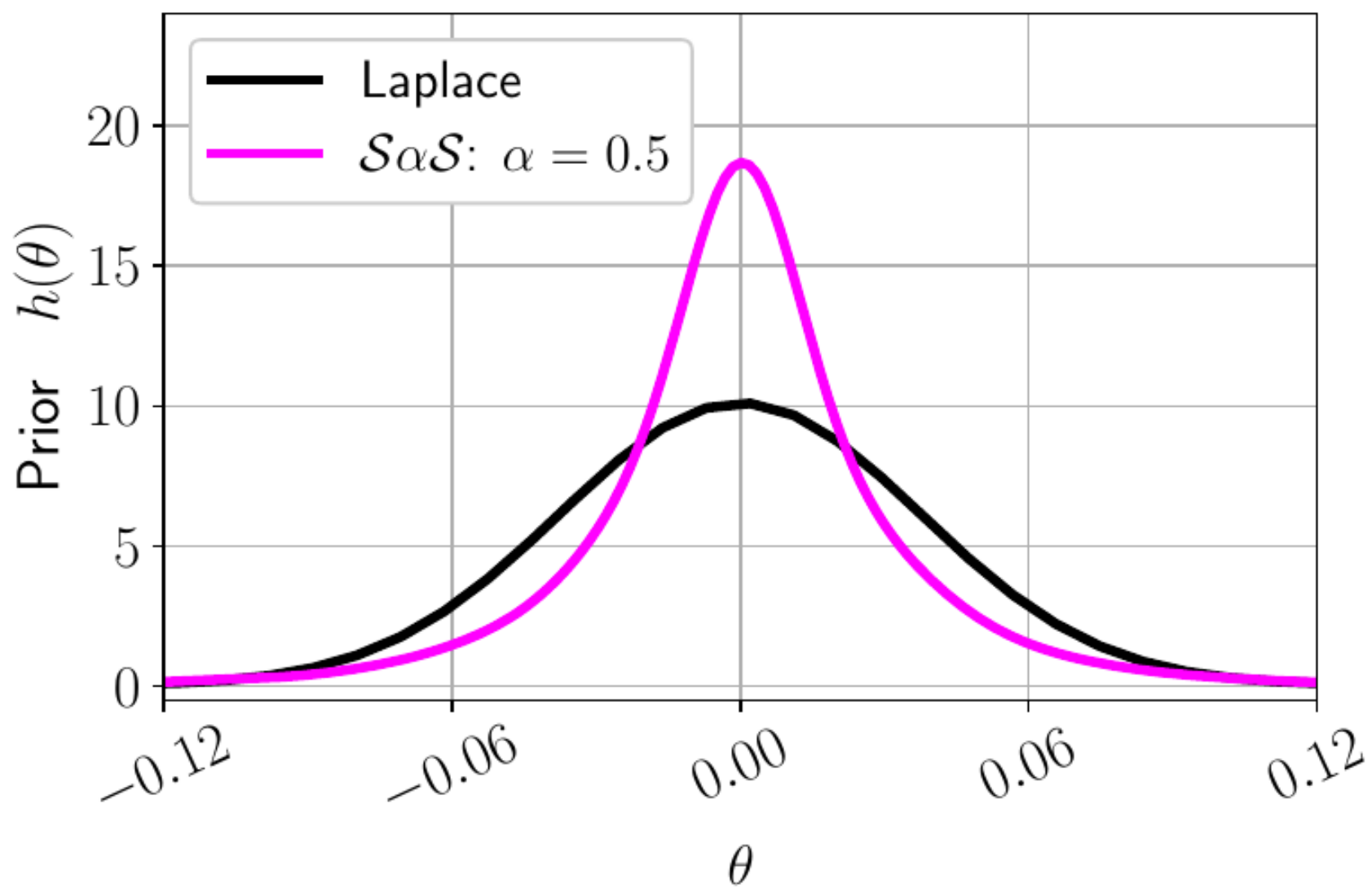

Figure 13: Sparsity comparison of Laplace (lasso) vs. soft diamond regularizers. The unimodal plots compare the post-training weight distribution of NMT models with non-uniform weight priors. The sub-Cauchy $\mathcal{S}\alpha\mathcal{S}$ prior with $\alpha = 0.5$ produced a sparser weight than the Laplacian prior because it concentrated far more weight values nearer zero than did the Laplacian prior.

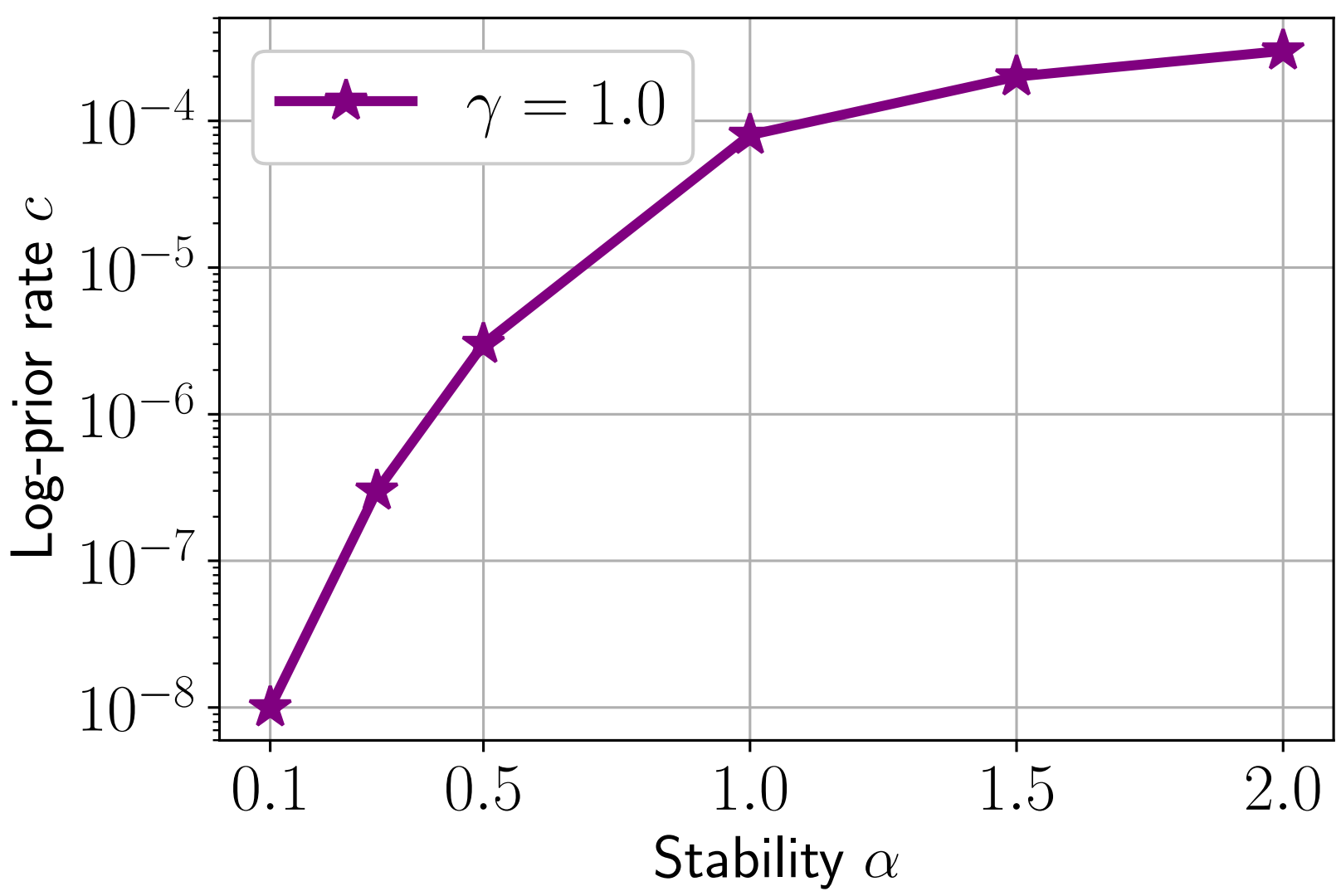

Figure 14: Best log-prior rate $c$ for $\mathcal{S}\alpha\mathcal{S}$ weight prior with neural machine translation models: The best log-prior rate increases as the stability value $\alpha$ of the $\mathcal{S}\alpha\mathcal{S}$ prior increases. We used a grid search to determine the optimal log-prior rates with respect to the BLEU metric.

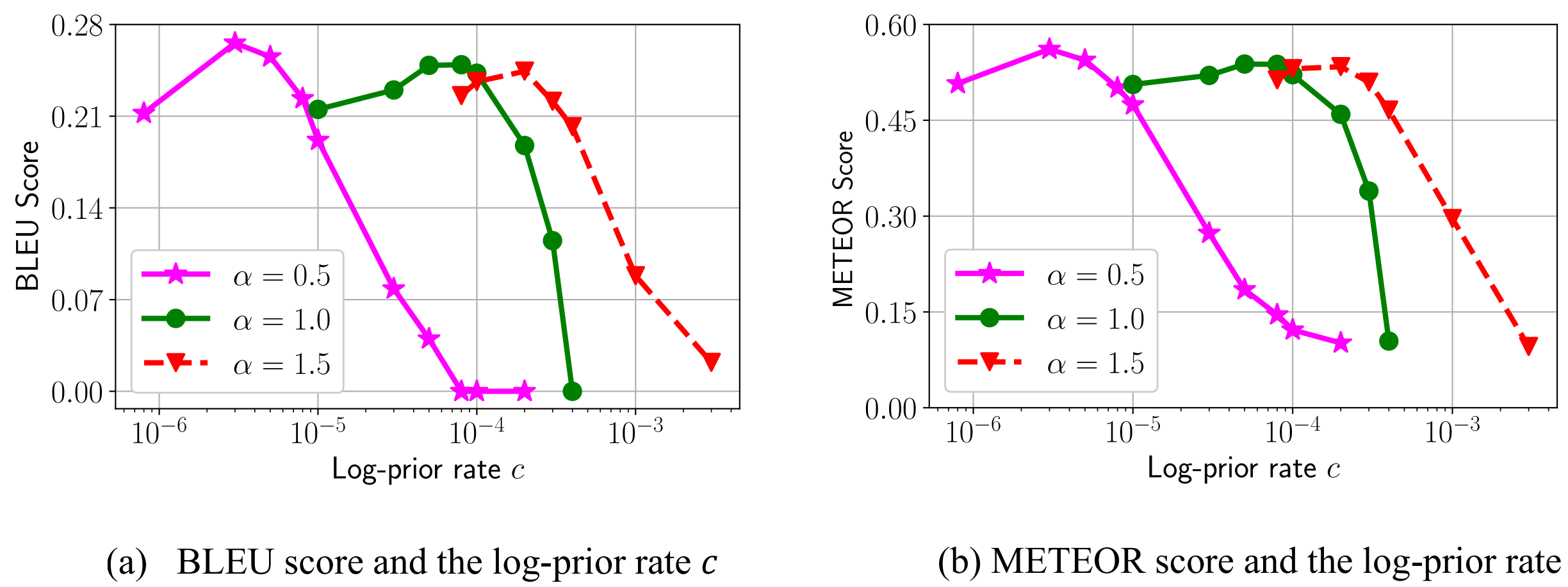

(a) BLEU score and the log-prior rate $c$

(b) METEOR score and the log-prior rate

Figure 15: Metrics vs. log-prior rate: This shows the effect of the log-prior rate for $\mathcal{S}\alpha\mathcal{S}$ on the performance of NMT models. We observed an inverted U-shaped relationship for the BLEU and METEOR metrics.

## Section V: Conclusion

Soft diamond regularizers produced a practical and efficient family of synaptic regularizers for deep networks compared with standard ridge (Gaussian) and lasso (Laplacian) regularizers. This new family of symmetric-$\alpha$-stable bell curves pose a computational problem because almost none of these continuum-many priors have a known closed-form mathematical description. We found that a simple precomputed table look-up overcame this problem and allowed straightforward computation. The popular Gaussian or $\ell_2$ regularizer had no sparsity at all: Almost all trained synaptic weights are nonzero. The best of the tested soft-diamond regularizers had better classification accuracy than the Gaussian regularizer. The Cauchy and $\alpha = 0.5$ regularizers had excellent weight sparsity and usually had the best classification among other soft diamond regularizers when training deep neural networks for classification or for regression.